\documentclass[journal]{IEEEtran}

\usepackage{comment}
\usepackage{multirow}
\usepackage{times}
\usepackage{epsfig}
\usepackage{graphicx}
\usepackage{amsmath}
\usepackage{amssymb}
\usepackage[dvipsnames]{xcolor}

\usepackage{booktabs}

\usepackage{cite}
\usepackage{hyperref}

\ifCLASSINFOpdf
\else
\fi

\usepackage{algorithmic}

\begin{document}
%

\title{Pretext-Contrastive Learning: Toward Good Practices in Self-supervised Video Representation Leaning}

%

\author{Li~Tao,~\IEEEmembership{Student~Member,~IEEE,}
        Xueting~Wang~\IEEEmembership{Member,~IEEE,}
        and~Toshihiko~Yamasaki,~\IEEEmembership{Member,~IEEE}} %

\markboth{Journal of \LaTeX\ Class Files,~Vol.~14, No.~8, August~2005}%
{Tao \MakeLowercase{\textit{et al.}}: Bare Demo of IEEEtran.cls for IEEE Journals}

\maketitle

\begin{abstract}
Recently, pretext-task based methods are proposed one after another in self-supervised video feature learning. Meanwhile, contrastive learning methods also yield good performance. Usually, new methods can beat previous ones as claimed that they could capture ``better'' temporal information. However, there exist setting differences among them and it is hard to conclude which is better. It would be much more convincing in comparison if these methods have reached as closer to their performance limits as possible. In this paper, we start from one pretext-task baseline, exploring how far it can go by combining it with contrastive learning, data pre-processing, and data augmentation. A proper setting has been found from extensive experiments, with which huge improvements over the baselines can be achieved, indicating a joint optimization framework can boost both pretext task and contrastive learning. We denote the joint optimization framework as Pretext-Contrastive Learning (PCL). The other two pretext task baselines are used to validate the effectiveness of PCL. And we can easily outperform current state-of-the-art methods in the same training manner, showing the effectiveness and the generality of our proposal. It is convenient to treat PCL as a standard training strategy and apply it to many other works in self-supervised video feature learning. 
Code will be made public at \url{https://github.com/BestJuly/Pretext-Contrastive-Learning}.
\end{abstract}


%
\IEEEpeerreviewmaketitle


\section{Introduction}
\label{sec:introduction}
\IEEEPARstart{W}ITH the development of convolutional neural networks (CNNs) and the help of many large-scale labeled datasets, the computer vision community has witnessed unprecedented success in many tasks such as object classification, detection, segmentation, and action recognition. For both image-level and video-level tasks, pre-training on larger datasets such as ImageNet~\cite{krizhevsky2012imagenet} and Kinetics~\cite{kinetics} is important to ensure satisfactory performance. 

However, the world is abundant in images and videos, and annotating large-scale datasets requires a wealth of resources. 
To leverage unlabeled data, many self-supervised learning methods have been proposed for efficient feature representation.
These methods can be broadly divided into two categories, pretext task-based methods and contrastive learning methods. 

\begin{figure}[t]
    \centering
    \includegraphics[width=\columnwidth]{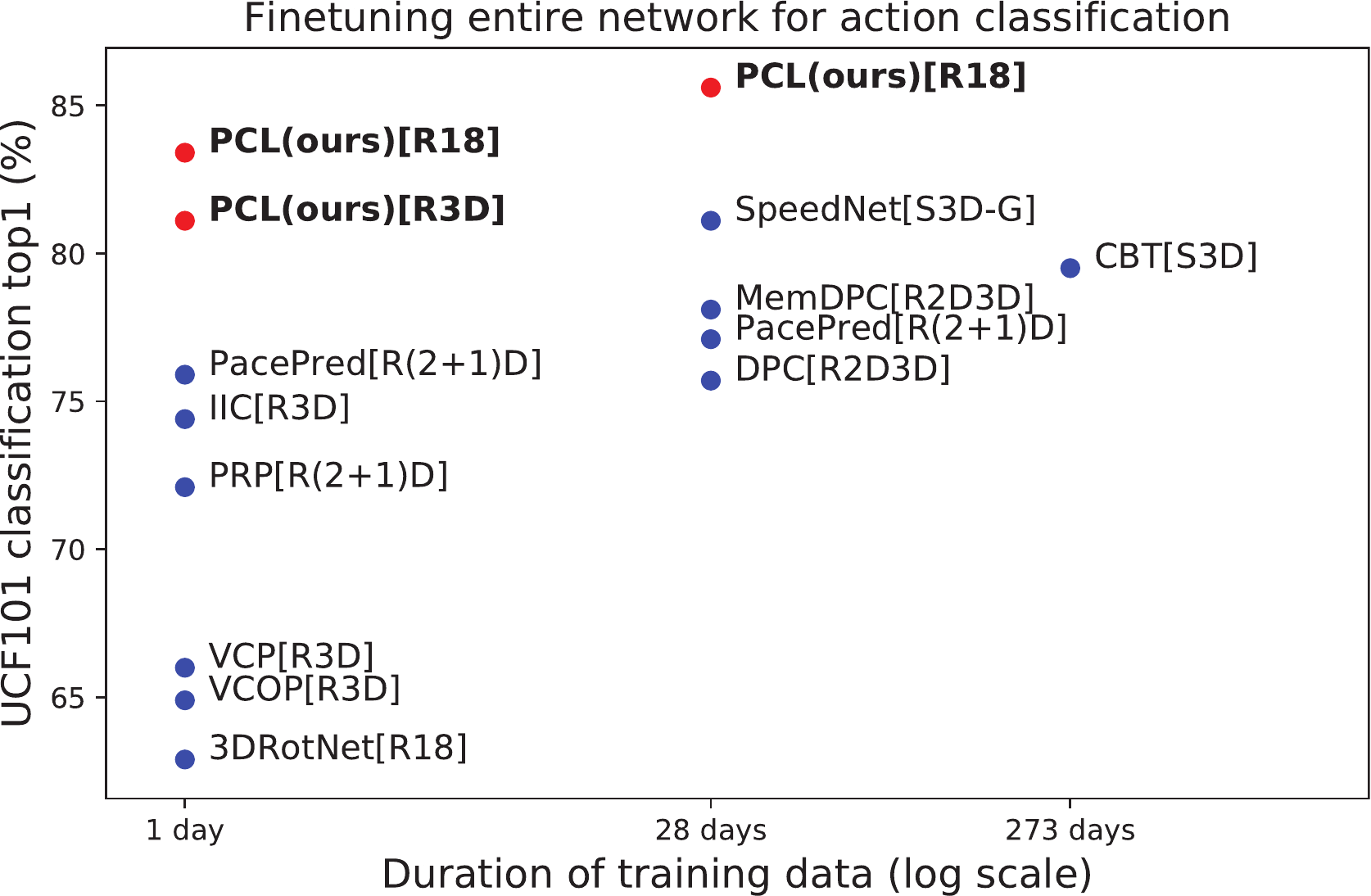} 
    \caption{A glance at the performance of our proposals. Our results in this figure are based on one pretext task, VCP~\cite{luo2020video}, the performance of which is only 66\%. Results of other methods are from corresponding papers and results using the same input sizes ($16\times112\times112$) are used if provided, without using other data modalities such as optical flow, audio and text.} 
    \label{fig:overview}
\end{figure}

Several tasks have been designed to constrain pretext task-based models to learn effective and informative representations. These tasks include solving jigsaw puzzles~\cite{noroozi2016unsupervised}, image inpainting~\cite{pathak2016context}, and detecting image rotation angles~\cite{gidaris2018unsupervised}. For video data, some of these spatial tasks are also effective, together with temporal-related tasks such as predicting frame orders~\cite{misra2016shuffle} or video clip orders~\cite{fernando2017self, lee2017unsupervised, xu2019self}, recognizing temporal transformations, and being sensitive to video playback speed~\cite{benaim2020speednet, cho2020self, yao2020video, jenni2020video, wang2020self}. A suitable combination~\cite{piergiovanni2020evolving} of such different tasks can help improve the performances of the methods in video retrieval and recognition tasks. 
However, even though high accuracy can be achieved, it seems to be endless because there can be new and ``better'' pretext tasks. Identifying which pretext task is more effective and why is theoretically difficult to explain.

In contrastive learning methods~\cite{chen2020simple,he2020momentum,wu2018unsupervised,oord2018representation,tian2019contrastive}, the solution is based on the comparison among different samples. The key idea is to distinguish one instance from another.
Usually, different modalities and different spatial/temporal crops of the same video are treated as positives while samples from different videos are treated as negatives, even though they may belong to the same action category. Once the network can distinguish one instance from another, the learned features would be sufficient for downstream tasks such as video retrieval and recognition.

The combination of different pretext tasks and contrastive learning seems to be better than each on its own. Such kind of combination using one pretext task (pace prediction) has been firstly validated effective in video representation learning reported in a recent work~\cite{wang2020self}. However, the reason why the combination can be effective lacks discussion and the generality of this combination is unsure whether this phenomenon happens only for a specific pretext task or not.

In this paper, based on the success of pretext tasks and contrastive learning, we want to explore what kind of combination can boost both. We propose Pretext-Contrastive Learning (PCL), which also facilitates the advantages of some data processing strategies such as residual clips~\cite{tao2020rethinking, tao2020motion} and strong data augmentations~\cite{chen2020simple}. With PCL, huge improvements over the corresponding baselines can be achieved, as shown in Fig.~\ref{fig:overview}. Better performance can be obtained over recent works while using much smaller (around 3.6\%) data for pre-training. We should clarify that we are not proposing new pretext tasks, nor contrastive learning methods; instead, we want to explore the limits of pretext task and contrastive learning with comprehensive experimental investigation and discussion to find the best strategy facilitating the advantages of these technologies. And this paper is trying to set new baselines in self-supervised learning in videos.


To prove the effectiveness of our PCL, three pretext task-based methods are set as baselines, together with the contrastive learning method. Different network backbones are tested to eliminate biases. Experimental results prove the effectiveness and the generality of our proposal. The proposed PCL is closer to a framework or a strategy rather than a simple method as it is flexible and can be applied to many existing solutions. And we have lifted benchmarks to a new level, setting new baselines in self-supervised video representation learning.

The contributions of this work can be summarized as:
\begin{itemize}
\item We propose a joint optimization framework, utilizing the advantage of both pretext tasks and contrastive learning, together with proper training settings.
\item Experiments demonstrate that huge improvements can be obtained by using our proposal, and we can also achieve state-of-the-art performances in two evaluation tasks on two benchmark datasets.
\item Our proposal is validated on the basis of three pretext task baselines and different network backbones, showing the effectiveness and the generality of our PCL.
\end{itemize}

\section{Related works}

\subsection{Pretext tasks}
Self-supervised learning methods were first proposed for images. Spatial pretext tasks include solving jigsaw puzzles~\cite{noroozi2016unsupervised}, detecting image rotations~\cite{gidaris2018unsupervised}, image channel prediction~\cite{zhang2016colorful}, and image inpainting~\cite{pathak2016context}. Prior works also include image reconstruction using autoencoders~\cite{hinton2006reducing} and variational autoencoders~\cite{kingma2013auto}.

For video data, some image-based pretext tasks can be directly applied or extended, such as detecting rotation angles~\cite{jing2018self} and completing space-time cubic puzzles~\cite{kim2019self}. Compared to image data, videos have an additional temporal dimension. Therefore, to utilize temporal information, many works have designed temporal-specific tasks. In~\cite{misra2016shuffle}, the network was trained to distinguish whether the input frames were in the correct order. \cite{fernando2017self} trained their odd-one-out network (O3N) to identify unrelated or odd video clips. The order prediction network (OPN)~\cite{lee2017unsupervised} was trained by predicting the correct order of shuffled frames. The video clip order prediction network~\cite{xu2019self} used video clips together with a spatio-temporal CNN during training. Further, \cite{luo2020video} utilized spatial and temporal transformations to train the network.
Many recent works 
have started to utilize the playback speed of the input video clips. SpeedNet~\cite{benaim2020speednet} was trained to detect whether a video is playing at a normal rate or a sped-up rate. \cite{cho2020self} trained a network to sort video clips according to the corresponding playback rates. The playback rate perception (PRP)~\cite{yao2020video} used an additional reconstructing decoder branch to help train the model. \cite{jenni2020video} and \cite{wang2020self} also utilized additional transformations to help train the model.

All these pretext tasks can be set as the main branch and can be combined with our PCL for better performance.

\subsection{Contrastive learning}
The success of contrastive learning also originated from image tasks~\cite{Jing2020pami}. The key idea of contrastive learning is to minimize the distance within positive pairs in the feature space while maximizing the distance between negative pairs. After contrastive loss was proposed~\cite{hadsell2006dimensionality}, contrastive learning has become the mainstream method for self-supervised learning of image data. Contrastive predictive coding~(CPC)~\cite{oord2018representation} attempted to learn the future from the past by using sequential data. Deep InfoMax~\cite{hjelm2018learning} and Instance Discrimination~\cite{wu2018unsupervised} were proposed to maximize information probability from the same sample. Contrastive multiview coding~(CMC)~\cite{tian2019contrastive} used different views (e.g. different color spaces) from the same sample. 
Momentum Contrast (MoCo)~\cite{he2020momentum, chen2020improved} used a momentum-updated encoder to conduct contrastive learning. In SimCLR~\cite{chen2020simple}, different combinations of data augmentation methods were tested for paired samples. Bootstrap Your Own Latent (BYOL)~\cite{grill2020bootstrap} trained the network without negative samples.

The above-mentioned methods mainly focus on image data. Some technologies have been successfully applied to video data. The concept of CMC can be easily adapted to videos by simply using video data as the model input. Similar to CPC, DPC~\cite{han2019video} and MemDPC~\cite{han2020memory} were proposed to handle video data. IIC~\cite{tao2020selfsupervised} introduced intra-negative video samples to enhance contrastive learning. These methods are all based on visual data only.
The contrastive learning concept can be extended to additional modalities of video, such as audio~\cite{owens2018audio, korbar2018cooperative}, text, and descriptive data~\cite{sun2019videobert}.

Most of these contrastive learning methods utilize a noise contrastive estimation (NCE) loss~\cite{gutmann2010noise} for robust and effective training. Wang et al.~\cite{wang2020hypersphere} explored the learned features and proposed a new loss function, align-uniform loss, which is a possible substitute for the NCE loss. In our PCL, we used the NCE loss for optimization. Other contrastive loss functions are also compatible in our framework.

\subsection{Methods combinations}
A combination of several pretext tasks with proper weights can yield better performances~\cite{piergiovanni2020evolving} than when they are used alone. Many existing pretext task-based methods are beyond one simple pretext task and are already a combination of some particular tasks. We have listed many pretext tasks, and the potential combinations are extensive. These pretext tasks vary widely, and determining why one pretext task or one combination is better than another is difficult.

The combination of pretext tasks and contrastive learning has been attempted in a recent work~\cite{wang2020self}. However, except for the reported results, few analyses have been conducted and the combination may be only effective on a specific task. In this paper, we address this issue and show the generality of the combination of pretext task and contrastive learning that it can boost performance of both. Improvements over three pretext task baselines also reveal that the effective settings can be generalized to a lot of pretext tasks.

\section{Methodology}

\subsection{Motivation}
Pretext task methods and contrastive learning methods can have good performances on their own. And the questions arise. 1) Can a simple combination of a pretext task based method and a contrastive learning method boost each other and achieve better performance? 2) Will it be effective only for specific pretext task, or general enough for many pretext tasks?

\subsection{PCL: pretext-contrastive learning}
The goal for self-supervised video representation learning is to learn effective feature representations from videos using a backbone network $f_\theta$. The commonly used networks are based on spatio-temporal convolutions, where the input video $v_i$ is decoded to a sequence of frames and several frames are stacked to form video clips $x_{v_i}$. Video features can be generated by using $f_\theta(x)$.

\textbf{Pretext task.} 
For pretext task-based methods, one or several tasks are used to train the network in a supervised manner. Most pretext tasks are classification tasks. For example, VCP~\cite{luo2020video} used different transformations on the input video clip $x$ and trained the network by distinguishing which transformation was conducted. 3DRotNet~\cite{jing2018self} was trained by detecting the rotation angles of the input clip. VCOP~\cite{xu2019self} shuffled video clips and trained the network by predicting the correct order class of the inputs. All these pretext tasks can be concluded as designing a proper classification task. The video clip~$x$ needs to be transformed by a specific transformation function~$t(x, y)$, where $y$ is the label of the corresponding transformation. Then the optimization target of these pretext tasks becomes

\begin{equation}
    \label{equ:pretext} 
    \underset{\forall v_i}{\text{minimize }} \mathcal{L}_{cls}(g(f_\theta(t(x_{v_i}, y))), y),
\end{equation}
where $g(\cdot)$ is the post-process network to process extracted features and $\mathcal{L}_{cls}$ is usually set as cross-entropy loss because the corresponding pretext tasks usually belong to classification tasks.

\begin{figure*}[t]
    \centering
    \includegraphics[width=1.8\columnwidth]{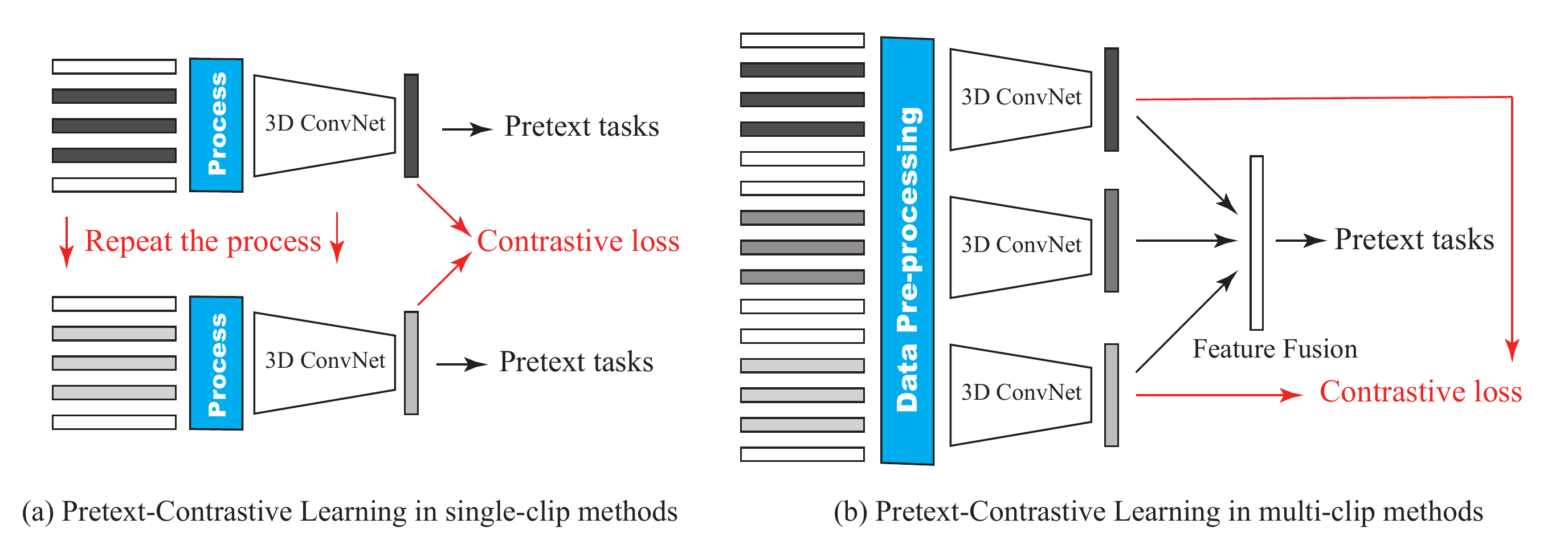} 
    \caption{The use of PCL in pretext task-based methods. (a) For single-clip methods, two different clips from the same video will be processed and the contrastive loss will be calculated among one batch of data. (b) For multi-clip methods, different clips from the same video have been already processed and the contrastive loss can be easily calculated. The data pre-processing procedure includes strong data augmentation transformations and converting to residual clips.} 
    \label{usage}
\end{figure*}

\textbf{Contrastive learning.}
For contrastive learning methods, after extracting features from the backbone, a two-linear-layer multi-layer perceptron (MLP) is usually used to project features $f_\theta(x)$ to another feature space. Let us denote the projector network as $h(\cdot)$. Positive pairs and negative pairs are required to constrain the network. $x_{v_i}^1$ is one video clip from the video $v_i$, and when another video clip $x_{v_i}^2$ is from the same video, these two video clips are treated as a positive pair. Conversely, when a video clip $x_{v_j}$ is from a different video, $v_j$. Then $x_{v_i}$ and $x_{v_j}$ are a negative pair. The encoded feature in the projected feature space is $h(f(x_{v_i}))$, which is denoted as $z_{v_i}$ for simplicity. Let us define $D(z_{v_i}, z_{v_j})$ as the similarity distance between feature $z_{v_i}$ and $z_{v_j}$; then for video $v_i$, the contrastive learning target is

\begin{equation}
    {\text{minimize }} \mathcal{L}_{NCE}^{v_i} = \mathcal{L}_{NCE}^{v_i^1} + \mathcal{L}_{NCE}^{v_i^2},
    \label{equ:nce} 
\end{equation}
where
\begin{equation}
    \label{equ:contrast} 
    \begin{aligned}
    \mathcal{L}_{NCE}^{v_i^1} = - \log\frac{D(z_{v_i}^1, z_{v_i}^2)}{D(z_{v_i}^1, z_{v_i}^2) + \sum_{j\neq{i}}{D(z_{v_i}^1, z_{v_j}^1)}},  \\
    \mathcal{L}_{NCE}^{v_i^2} = - \log\frac{D(z_{v_i}^1, z_{v_i}^2)}{D(z_{v_i}^1, z_{v_i}^2) + \sum_{j\neq{i}}{D(z_{v_i}^2, z_{v_j}^2)}}.
    \end{aligned}
\end{equation}
In practice, video features (i.e., $z_{v_i}$) are normalized in the feature space and the similarity distance $D(z_{v_i}, z_{v_j})$ is calculated by the inner product. In contrastive learning, instances with different indexes can be treated as negative samples and at most $N-1$ negative samples can be used, where $N$ is the size of the dataset. To accelerate training, memory bank~\cite{wu2018unsupervised} technologies are adopted to save extracted features from previous epochs and $k$ negative samples are sampled from the corresponding memory banks. This procedure is similar to~\cite{tian2019contrastive,tao2020selfsupervised}.

\textbf{Joint optimization framework.}
As we can see from the optimization targets of pretext tasks (Eq.~\ref{equ:pretext}) and contrastive learning (Eq.~\ref{equ:nce} and Eq.~\ref{equ:contrast}), pretext task-based methods focus more within the sample while contrastive learning methods try to distinguish one instance from another. A combination of them may take the advantage of both, ensuring the network to have a local-global view. 

There are a number of pretext tasks, and some tasks use only one video clip to conduct experiments such as 3DRotNet~\cite{jing2018self}, which rotated the input video clip and trained the model by predicting the rotation angles. Some tasks use multiple video clips during training, such as VCOP~\cite{xu2019self}, which shuffled the temporal order of several video clips. The training styles for almost all pretext tasks can be divided into two categories, single-clip methods and multi-clip methods.

We illustrate the use of our proposal in Fig.~\ref{usage}. For single-clip methods, the contrastive loss will use the encoded features from the backbone network. As contrastive loss requires positive pair and negative pairs to train, the encoding process is duplicated. The input video clip is generated from the same video as the original path, which can be treated as a positive pair. Negative pairs are taken directly from one batch of data because different samples are from different videos in one training batch.

For multi-clip methods, different video clips are set as inputs and they are encoded to features using a shared encoder. These features are natural positive pairs because they are from the same video. Negative pairs are also from video clips from one batch data.

It can be observed that it is very simple to construct a joint optimization framework based on any pretext task baseline method, and the final training loss becomes

\begin{align}
    \mathcal{L}_{total} & = \mathcal{L}_{pretext} + \alpha \mathcal{L}_{contrast},
    \label{eq:loss}
\end{align}
where $\alpha$ is a weight to balance losses between pretext tasks and contrastive learning. For $\mathcal{L}_{pretext}$ and $\mathcal{L}_{contrast}$, they came from Eq.~\ref{equ:pretext} and Eq.~\ref{equ:contrast}. For convenience, we rewrite them here.
\begin{align}
    \mathcal{L}_{pretext} & = \mathcal{L}_{cls}(g(f_\theta(t(x_{v_i}, y))), y), \\
    \mathcal{L}_{contrast} & = \mathcal{L}_{NCE}^{v_i^1} + \mathcal{L}_{NCE}^{v_i^2}
    \label{eq:detailed_loss}
\end{align}

\subsection{Data processing strategies}
\label{subsec:preprocess}
To further boost the performance, we mainly introduce two different kinds of processing strategies on data, residual clips and augmentation transformations. 

\textbf{Residual clips.}
Most video-based self-supervised learning methods use 3D convolutional networks to process data, and the corresponding input is video clips, which are stacked RGB frames. Residual clips were introduced in~\cite{tao2020rethinking} and have been used to self-supervised learning in ~\cite{tao2020selfsupervised}. However, there are not many researches on whether it also functions well with several pretext tasks, nor its generality. We introduce residual clips here to show its effectiveness on different methods in self-supervised learning.

Here we use $frame_i$ to represent the $i_{th}$ frame data, and $Frame_{i\sim j}$ denotes the stacked frames from the $i_{th}$ frame to the $j_{th}$ frame. The process to get residual frames can be formulated as follows,

\begin{equation}
   ResClip = Frame_{i+1\sim j+1} - Frame_{i\sim j},
\end{equation}
where $Frame_{i+1\sim j+1}$ can be easily obtained by shifting frames along the temporal axis in video clips.

The $ResClip$ here will be then directly fed into the network for feature extraction.

\textbf{Augmentation transformations.}
It is widely acknowledged that data augmentation methods enhance the performance in most cases. However, in previous methods in video-based self-supervised learning, only few data augmentations were conducted such as random cropping in the spatial axis and temporal jittering. However, some recent works~\cite{he2020momentum, grill2020bootstrap} started to use strong augmentations in images, such as color distortion and Gaussian blur, and have achieved improvements over the corresponding baselines.

Though these data augmentations are conducted on images, we adopted this kind of processing and applied it to video frames. The motivation is that motion features should be similar even though frames are blurred or distorted by color. Another motivation is that we wonder whether it will also boost the performance of residual clips because there will be much fewer color information in
residual clips.

\section{Experiments}
To demonstrate the effectiveness of the proposed PCL framework, we used 1) three pretext task baselines and a contrastive learning baseline method, 2) four backbone networks, and 3) two evaluation tasks in our experiments.

\subsection{Data}
We mainly used two benchmark datasets, UCF101~\cite{ucf101} and HMDB51~\cite{hmdb} in our experiments, as our baselines did~\cite{jing2018self, xu2019self, luo2020video}. The UCF101 dataset consists of 13,320 videos in 101 action categories. HMDB51 is comprised of 7,000 videos with a total of 51 action classes. The official splits only contain training set and testing set. We randomly selected 800 and 400 videos from the training splits for UCF101 and HMDB51 datasets, respectively, to form the validation set. The best performance on the validation set will be saved and evaluated in video retrieval and recognition. To further evaluate the effectiveness of our PCL, we also utilized Kinetics-400 dataset~\cite{i3d} to train. Kinetics-400 consists 400
action classes and contains around 240k videos for training, 20k videos for validation and 40k videos for testing. We only used Kinetics-400 dataset in the pre-training process.

Because spatio-temporal convolutions were used to train our models, we followed~\cite{c3d} and resized videos in size $128\times171$. Sixteen successive frames are sampled to form a video clip. Random spatial/temporal cropping was conducted to generate an input data of size $16\times112\times112$, where the channel number $3$ was ignored. In addition to random cropping, other augmentation transformations we used in our experiments include random color jittering, randomly converting to grayscale, Gaussian blur, and flipping.

\subsection{Baselines}
Because our PCL is a combination of pretext tasks and contrastive learning, the baselines should be set as the pretext task or contrastive learning. 

\begin{table*}[tb]
    \centering
      \caption{Comparisons with baselines. Results are evaluated on \textit{split}~1 of UCF101 and HMDB51. \textit{Contrastive only} represents we only use contrastive learning loss only in Eq.~\ref{eq:loss} for network optimization, which already benefits from our data processing strategies which have been introduced in Sec.~\ref{subsec:preprocess}.
      Best results in each block are in \textbf{bold}. }
      \scalebox{1.0}{
      \renewcommand{\arraystretch}{1.1}
      \setlength{\tabcolsep}{1.2mm}{
      \begin{tabular}{c c  c c c c c  c  c c c c c  c }
        \toprule
        \multirow{2}{*}{Method} & \multirow{2}{*}{Backbone} & \multicolumn{6}{c}{UCF101} & \multicolumn{6}{c}{HMDB51}
        \\
        \cmidrule(lr){3-8} \cmidrule(lr){9-14}
        &   & Top1 & Top5 & Top10 & Top20 & Top50 & Recognition & Top1 & Top5 & Top10 & Top20 & Top50 & Recognition\\
        \midrule
        VCP (baseline)~\cite{luo2020video} & C3D & 17.3  & 31.5  & 42.0  & 52.6  & 67.7  & 68.5  & 7.8  & 23.8  & 35.3  & 49.3  & 71.6  & 32.5 \\
        Contrastive only & C3D & 38.9 & 56.9 & 65.7 & 74.4 & 84.3 & 78.0 & 15.1 & 34.9 & 47.2 & 61.5 & 82.1 & 45.5\\
        PCL & C3D & \textbf{50.3} & \textbf{67.3} & \textbf{75.7} & \textbf{83.4} & \textbf{91.2} & \textbf{79.8} & \textbf{19.6} & \textbf{41.5} & \textbf{44.8} & \textbf{70.2} & \textbf{85.9} & \textbf{46.1} \\
        \hline
        VCP (baseline)~\cite{luo2020video} & R3D & 18.6  & 33.6  & 42.5  & 53.5  & 68.1  & 66.0  & 7.6  & 24.4  & 36.3  & 53.6  & 76.4  & 31.5 \\
        Contrastive only & R3D & 44.7 & 62.4 & 71.6 & 79.6 & 88.8 & 79.3 & 17.3 & 38.6 & 51.2 & 65.3 & 83.4 & \textbf{46.3} \\
        PCL & R3D & \textbf{48.1} & \textbf{64.7} & \textbf{73.9} & \textbf{82.0} & \textbf{90.6} & \textbf{79.9} & \textbf{19.2} & \textbf{42.0} & \textbf{55.3} & \textbf{69.1} & \textbf{86.7} & 46.1 \\
        \hline
        VCP (baseline)~\cite{luo2020video} & R(2+1)D & 19.9  & 33.7  & 42.0  & 50.5  & 64.4  & 66.3  & 6.7  & 21.3  & 32.7  & 49.2  & 73.3  & 32.2 \\
        PCL & R(2+1)D & \textbf{42.8} & \textbf{59.9} & \textbf{69.5} & \textbf{78.0} & \textbf{87.6} & \textbf{79.9} & \textbf{19.6} & \textbf{41.1} & \textbf{56.2} & \textbf{71.1} & \textbf{86.5} & \textbf{45.9} \\
        \midrule
        3DRotNet (baseline)~\cite{jing2018self} & R3D-18 & 14.2  & 25.2  & 33.5  & 43.7  & 59.5  & 62.9  & 6.2  & 18.7  & 31.0  & 46.6  & 70.5  & 33.7 \\
        PCL & R3D-18 & \textbf{33.7} & \textbf{53.5} & \textbf{64.1} & \textbf{73.4} & \textbf{85.0} & \textbf{81.5} & \textbf{12.4} & \textbf{34.4} & \textbf{48.4} & \textbf{65.4} & \textbf{83.6} & \textbf{47.4} \\
        \midrule
        VCOP (baseline)~\cite{xu2019self} & C3D & 12.5  & 29.0  & 39.0  & 50.6  & 66.9  & 65.6  & 7.4  & 22.6  & 34.4  & 48.5  & 70.1  & 28.4 \\
        PCL & C3D & \textbf{39.0} & \textbf{59.1} & \textbf{67.5} & \textbf{76.8} & \textbf{87.4} & \textbf{79.2} & \textbf{14.9} & \textbf{35.9} & \textbf{48.9} & \textbf{63.6} & \textbf{82.8} & \textbf{42.2} \\
        \bottomrule

      \end{tabular}}}
      \label{table:compare_baseline}
    \end{table*}

There are several pretext task-based methods in self-supervised video representation learning. We chose three works:  3DRotNet~\cite{jing2018self}, VCOP~\cite{xu2019self}, and VCP~\cite{luo2020video}. 3DRotNet is trained by recognizing the rotated angles of the input video clip. VCOP aims to detect the correct orders of several input video clips. VCP conducts different types of transformations and the network is trained to distinguish which transformation has been performed. These pretext tasks, as well as the training styles, are different. For example, 3DRotNet is a one-clip method while VCOP and VCP use several video clips as input data. The other reason is that these three pretext tasks are from three different categories. 3DRotNet uses rotation, which is more related to spatial information. VCOP cares about temporal orders of input clips, which only uses temporal information. The processing which VCP chooses from is a mixture of temporal and spatial transformations. There exist many pretext tasks in video-based self-supervised learning and it is impossible for us to conduct all experiments. However, other pretext tasks can be easily classified into one of these three categories and we think the effectiveness of our proposal on these three pretext tasks can prove the generality of our PCL.

Contrastive learning is widely used in image-based self-supervised learning and has been explored in videos in~\cite{tao2020selfsupervised, han2020memory, han2019video}. For a fair comparison, our contrastive learning baseline will use the same framework as our PCL while the network will be optimized only by $\mathcal{L}_{contrast}$ in Eq.~\ref{eq:loss}, without using $\mathcal{L}_{pretext}$.

\subsection{Network backbones}
For the network backbone, there are several 3D CNNs such as C3D~\cite{c3d}, R3D, ResNet-18-3D~\cite{res3d}, and R(2+1)D~\cite{r3d}. Different network backbones were used in our experiments to eliminate model biases. R3D and ResNet-18-3D are composed of 3D convolution instead of 2D convolution in the original ResNet~\cite{resnet} while the numbers of convolutional layers in each residual block vary. To compare with the baselines, we used the same network architectures as them. It is possible to use other network architectures such as I3D~\cite{i3d}, S3D~\cite{s3d}, or other deeper networks, but we simply follow the baselines for fair comparisons.

A two-linear-layer multi-layer perceptron (MLP) is used to process features from the same backbone. Therefore, this part can be treated as the post-processing for the contrastive learning part, paralleling with the post-processing of pretext tasks. The MLP is in an \textit{fc-relu-fc} style. After projection, feature dimensions are reduced to 128 in our experiments.

\subsection{Evaluation tasks}
To evaluate the performance of the trained models, two evaluation tasks were used: video retrieval and video recognition. After self-supervised training, the trained models can be evaluated directly in video retrieval tasks on both UCF101 and HMDB51. Note that the self-supervised learning part was only conducted on UCF101 \textit{split}~1. Therefore, when conducting video retrieval on UCF101, the task-level generalization ability was tested. When conducting video retrieval on HMDB51 using the same model, both task-level and dataset-level generalization abilities were tested.

Video retrieval is conducted based on video-level features. 3D ConvNets can extract features from video clips, and features of video clips are averaged if they are from the same video. Thus, video-level features can be generated and k-nearest neighbors (kNN) algorithm is used to check whether the retrieved video has the same action category as the query video.

Action recognition is a fundamental task in video representation learning. Following previous works, we also conducted experiments by fine-tuning trained models on both UCF101 and HMDB51 datasets to check the transfer learning ability of the models.

\subsection{Experimental details}
In all of our experiments, the batch size is set to 16 and the training lasts for 200 epochs. The initial learning rate is 0.01 for self-supervised learning. Models with the best performance on the validation datasets are saved then used to test the performance in the video retrieval task. For video recognition tasks, the same models are fine-tuned for 150 epochs and the initial learning rate is set to 0.001. The best performance on the validation dataset is evaluated on the corresponding test splits. Stochastic Gradient Descent (SGD) is used for optimization for both training periods. The hyper-parameter in Eq.~\ref{eq:loss}, $\alpha$ is set to 0.5 to balance pretext task loss and contrastive loss.

\section{Experimental Results}
In this section, we first compare our proposed method with baseline methods. To further prove the effectiveness of our PCL framework, we also compare our results with current state-of-the-art methods. We mainly used VCP as the baseline pretext task and used C3D, R3D, or R(2+1)D as the network backbone. For the other two methods, 3DRotNet and VCOP, we used the same mainstream backbones reported in the corresponding papers: ResNet-18-3D for 3DRotNet and C3D for VCOP.

\subsection{Comparison with baselines}
All models were pre-trained on UCF101 \textit{split}~1 and tested on both UCF101 and HMDB51 datasets. Results are presented in Table~\ref{table:compare_baseline}.

For the pretext task VCP with the C3D backbone, the baseline is only 17.3\% in video retrieval and 68.5\% in recognition on UCF101 dataset. When maintaining the main training architecture and used contrastive loss only, the performance can reach 38.9\% in retrieval and 78.0\% in video recognition. This performance is much higher than the pretext task. One reason is that it already benefits from our data processing strategies. 
Our PCL yielded 50.3\% at top~1 retrieval accuracy on UCF101 dataset, which is 33.0\% points above the C3D baseline for pretext task and also 11.4\% points higher than our strong contrastive learning baseline. In video recognition, our PCL can also yield the best performance. 

Similar results can be found when we used different network backbones on the basis of VCP. Our PCL can achieve more than double the performance of the corresponding baseline at top~1 retrieval accuracy and over 10\% points improvement when we use R3D and R(2+1)D as the network backbone. These results show that our PCL can boost the performance of both VCP and contrastive learning. 

When we look at other pretext baselines in Table~\ref{table:compare_baseline}, the same trend can be found. Our PCL can outperform the corresponding pretext task baselines, 3DRotNet and VCOP, by a large margin. These results reveal that the effectiveness of our PCL is not limited to only one pretext tasks, but general enough to other methods. Also, we want to mention that VCP cares much about spatial and temporal transformations, VCOP uses temporal information only and 3DRotNet uses rotation which is much more related to spatial information. The effectiveness of PCL on these three baselines reveals the potential that our PCL can boost the performance of other existing pretext task based methods in self-supervised video representation learning. 

\begin{table}[tb]
  \centering
    \caption{Comparison with state-of-the-art methods in video retrieval on UCF101. Most results are from the corresponding papers.}
    \scalebox{1.1}{
    \setlength{\tabcolsep}{0.5mm}{
    \begin{tabular}{c c c c c c c}
        \toprule
        Methods  & Backbone & Top1 & Top5 & Top10 & Top20 & Top50\\
        \midrule
        MemDPC~\cite{han2020memory} & R2D3D & 20.2 & 40.4 & 52.4 & 64.7 & - \\
        MemDPC-Flow~\cite{han2020memory} & R2D3D & \textbf{40.2} & \textbf{63.2} & \textbf{71.9} & \textbf{78.6} & - \\
        \midrule
        \textit{Random} & C3D & 16.7 & 27.5 & 33.7 & 41.4 & 53.0 \\
        VCOP~\cite{xu2019self} & C3D & 12.5 & 29.0 & 39.0 &	50.6 & 66.9 \\
        VCP~\cite{luo2020video} & C3D & 17.3 & 31.5 & 42.0 & 52.6 & 67.7 \\
        PRP~\cite{yao2020video} & C3D & 23.2 & 38.1 & 46.0 & 55.7 & 68.4 \\
        PacePred~\cite{wang2020self} & C3D & 31.9 & 49.7 & 59.2 & 68.9 & 80.2 \\
        IIC~\cite{tao2020selfsupervised} & C3D & 31.9 & 48.2 & 57.3 & 67.1 & 79.1 \\
        \textbf{PCL (VCOP)} & C3D & 39.0 &	59.1 &	67.5 &	76.8 &	87.4 \\
        \textbf{PCL (VCP)} & C3D & \textbf{50.3} &	\textbf{67.3} &	\textbf{75.7} &	\textbf{83.4} &	\textbf{91.2} \\
        \midrule
        \textit{Random} & R3D & 9.9 & 18.9 & 26.0 & 35.5 & 51.9 \\
        VCOP~\cite{xu2019self} & R3D & 14.1 & 30.3 & 40.0 & 51.1 & 66.5 \\
        VCP~\cite{luo2020video} & R3D & 18.6 & 33.6 & 42.5 & 53.5 & 68.1 \\
        PRP~\cite{yao2020video} & R3D & 22.8 & 38.5 & 46.7 & 55.2 & 69.1 \\
        IIC~\cite{tao2020selfsupervised} & R3D & 36.5 & 54.1 & 62.9 & 72.4 & 83.4 \\
        \textbf{PCL (VCOP)} & R3D & 38.9 & 57.8 & 66.6 & 76.1 & 86.0	\\
        \textbf{PCL (VCP)} & R3D & \textbf{48.1} & \textbf{64.7} & \textbf{73.9} & \textbf{82.0} & \textbf{90.6} \\
        \midrule
        \textit{Random} & R(2+1)D & 10.6 & 20.7 & 27.4 & 37.4 & 53.1 \\
        VCOP~\cite{xu2019self} & R(2+1)D & 10.7 & 25.9 & 35.4 & 47.3 & 63.9 \\
        VCP~\cite{luo2020video} & R(2+1)D & 19.9 & 33.7 & 42.0 & 50.5 & 64.4 \\
        PRP~\cite{yao2020video} & R(2+1)D & 20.3 & 34.0 & 41.9 & 51.7 & 64.2 \\
        PacePred~\cite{wang2020self} & R(2+1)D & 25.6 & 42.7 & 51.3 & 61.3 & 74.0 \\
        IIC~\cite{tao2020selfsupervised} & R(2+1)D & 34.7 & 51.7 & 60.9 & 69.4 & 81.9 \\
        \textbf{PCL (VCOP)} & R(2+1)D & 16.6 & 33.3 & 43.1 & 55.5 & 72.6 \\
        \textbf{PCL (VCP)} & R(2+1)D & \textbf{42.8} & \textbf{59.9} & \textbf{69.5} & \textbf{78.0} & \textbf{87.6} \\
        \midrule
        \textit{Random} & R3D-18 & 15.3 & 25.1 & 32.1 & 40.8 & 53.7 \\
        3DRotNet & R3D-18 & 14.2 & 25.2 & 33.5 & 43.7 & 59.5 \\
        VCP~\cite{luo2020video} & R3D-18 & 22.1 & 33.8 & 42.0 & 51.3 & 64.7 \\
        RTT~\cite{jenni2020video} & R3D-18 & 26.1 & 48.5 & 59.1 & 69.6 & 82.8 \\
        PacePred~\cite{wang2020self} & R3D-18 & 23.8 & 38.1 & 46.4 & 56.6 & 69.8 \\
        IIC~\cite{tao2020selfsupervised} & R3D-18 & 36.8 & 54.1 & 63.1 & 72.0 & 83.3 \\
        \textbf{PCL (3DRotNet)} & R3D-18 & 33.7 & 53.5 & 64.1 & 73.4 & 85.0 \\
        \textbf{PCL (VCP)} & R3D-18 & \textbf{55.1} & \textbf{71.2} & \textbf{78.9} & \textbf{85.5} & \textbf{92.3} \\
        \bottomrule
    \end{tabular}}}
    \label{table:retrieval_sota}
  \end{table}

\begin{table}[t]
  \centering
    \caption{Comparison with state-of-the-art methods in video retrieval on HMDB51. Most results are from the corresponding papers.}
    \scalebox{1.1}{
    \setlength{\tabcolsep}{0.5mm}{
    \begin{tabular}{c c c c c c c}
        \toprule
        Methods  & Backbone & Top1 & Top5 & Top10 & Top20 & Top50\\
        \midrule
        MemDPC~\cite{han2020memory} & R2D3D & 7.7 & 25.7 & 40.6 & 57.7 & - \\
        MemDPC-Flow~\cite{han2020memory} & R2D3D & \textbf{15.6} & \textbf{37.6} & \textbf{52.0} & \textbf{65.3} & - \\
        \midrule
        \textit{Random} & C3D & 7.4 & 20.5 & 31.9 & 44.5 & 66.3 \\
        VCOP~\cite{xu2019self} & C3D & 7.4 & 22.6 & 34.4 & 48.5 & 70.1 \\
        VCP~\cite{luo2020video} & C3D & 7.8 & 23.8 & 35.3 & 49.3 & 71.6 \\
        PRP~\cite{yao2020video} & C3D & 10.5 & 27.2 & 40.4 & 56.2 & 75.9 \\
        PacePred~\cite{wang2020self} & C3D & 12.5 & 32.2 & 45.4 & 61.0 & 80.7 \\
        IIC~\cite{tao2020selfsupervised} & C3D & 11.5 & 31.3 & 43.9 & 60.1 & 80.3 \\
        \textbf{PCL (VCOP)} & C3D & 14.9 & 35.9 & 48.9 & 63.6 & 82.8 \\
        \textbf{PCL (VCP)} & C3D & \textbf{19.6} & \textbf{41.5} & \textbf{44.8} & \textbf{70.2} & \textbf{85.9} \\
        \midrule
        \textit{Random} & R3D & 6.7 & 18.3 & 28.3 & 43.1 & 67.9 \\
        VCOP~\cite{xu2019self} & R3D & 7.6 & 22.9 & 34.4 & 48.8 & 68.9 \\
        VCP~\cite{luo2020video} & R3D & 7.6 & 24.4 & 36.3 & 53.6 & 76.4 \\
        PRP~\cite{yao2020video} & R3D & 8.2 & 25.8 & 38.5 & 63.3 & 75.9 \\
        IIC~\cite{tao2020selfsupervised} & R3D & 13.4 & 32.7 & 46.7 & 61.5 & 83.8 \\
        \textbf{PCL (VCOP)} & R3D & 14.3 & 34.0 & 48.3 & 62.1 & 81.9\\
        \textbf{PCL (VCP)} & R3D & \textbf{19.2} & \textbf{42.0} & \textbf{55.3} & \textbf{69.1} & \textbf{86.7}\\
        \midrule
        \textit{Random} & R(2+1)D & 4.5 & 14.8 & 23.4 & 38.9 & 63.0 \\
        VCOP~\cite{xu2019self} & R(2+1)D & 5.7 & 19.5 & 30.7 & 45.6 & 67.0 \\
        VCP~\cite{luo2020video} & R(2+1)D & 6.7 & 21.3 & 32.7 & 49.2 & 73.3 \\
        PRP~\cite{yao2020video} & R(2+1)D & 8.2 & 25.3 & 36.2 & 51.0 & 73.0 \\
        PacePred~\cite{wang2020self} & R(2+1)D & 12.9 & 31.6 & 43.2 & 58.0 & 77.1 \\
        IIC~\cite{tao2020selfsupervised} & R(2+1)D & 12.7 & 33.3 & 45.8 & 61.6 & 81.3 \\
        \textbf{PCL (VCOP)} & R(2+1)D & 7.9 & 23.8 & 35.9 & 51.0 & 74.7 \\
        \textbf{PCL (VCP)} & R(2+1)D & \textbf{19.6} & \textbf{41.1} & \textbf{56.2} & \textbf{71.1} & \textbf{86.5} \\
        \midrule
        \textit{Random} & R3D-18 & 7.1 & 19.3 & 29.6 & 44.2 & 68.6 \\
        3DRotNet & R3D-18 & 6.2 & 18.7 & 31.0 & 46.6 & 70.5 \\
        VCP~\cite{luo2020video} & R3D-18 & 10.9 & 25.2 & 36.8 & 51.5 & 71.8 \\
        PacePred~\cite{wang2020self} & R3D-18 & 9.6 & 26.9 & 41.1 & 56.1 & 76.5 \\
        IIC~\cite{tao2020selfsupervised} & R3D-18 & 15.5 & 34.4 & 48.9 & 63.8 & 83.8 \\
        \textbf{PCL (3DRotNet)} & R3D-18 & 12.4 & 34.4 & 48.4 & 65.4 & 83.6 \\
        \textbf{PCL (VCP)} & R3D-18 & \textbf{20.2} & \textbf{43.6} & \textbf{59.1} & \textbf{72.5} & \textbf{86.6} \\
        \bottomrule
    \end{tabular}}}
    \label{table:retrieval_hmdb}
  \end{table}

\setlength{\tabcolsep}{2pt}
\begin{table}[t]
\footnotesize
	\centering
	\caption{Comparisons with the state-of-the-art self-supervised methods in video recognition on UCF101 and HMDB51 dataset (pre-trained on video modality only, without using optical flow data). \textit{PCL (VCP)} represent our PCL which is based on the pretext task VCP. And results from our PCL are reported with proposed data processing strategies such as using residual clips and data augmentations.}
		\begin{tabular}{lcccccc} 
		\toprule
		Method &  Date & Pre-train & ClipSize & Network & UCF  & HMDB  \\ 
		\midrule
		OPN~\cite{lee2017unsupervised}      & 2017   & UCF    & $227^2$               & VGG     & 59.6 & 23.8  \\
		DPC~\cite{han2019video}             & 2019   & K400   & $16\times224^2$     & R3D-34  & 75.7 & 35.7  \\
		CBT~\cite{sun2019videobert}         & 2019   & K600+  & $16\times112^2$     & S3D     & 79.5 & 44.6  \\
		SpeedNet~\cite{benaim2020speednet}  & 2020   & K400   & $64\times224^2$     & S3D-G   & 81.1 & 48.8  \\
		MemDPC~\cite{han2020memory}         & 2020   & K400   & $40\times224^2$     & R-2D3D  & 78.1 & 41.2  \\
		\midrule
		VCOP~\cite{xu2019self}              & 2019   & UCF    & $16\times112^2$     & C3D     & 65.6 & 28.4  \\
		VCP~\cite{luo2020video}             & 2020   & UCF    & $16\times112^2$     & C3D     & 68.5 & 32.5  \\
		PRP~\cite{yao2020video}             & 2020   & UCF    & $16\times112^2$     & C3D     & 69.1 & 34.5  \\
		RTT~\cite{jenni2020video}           & 2020   & K400   & $16\times112^2$     & C3D     & 69.9 & 39.6  \\
		\textbf{PCL (VCOP)} & & UCF        & $16\times112^2$     & C3D     & 79.8 & 41.8  \\
		\textbf{PCL (VCP)} & & UCF         & $16\times112^2$     & C3D     & \textbf{81.4} & \textbf{45.2}  \\
		\midrule
		VCOP~\cite{xu2019self}              & 2019   & UCF    & $16\times112^2$     & R3D     & 64.9 & 29.5  \\
		VCP~\cite{luo2020video}             & 2020   & UCF    & $16\times112^2$     & R3D     & 66.0 & 31.5  \\
		PRP~\cite{yao2020video}             & 2020   & UCF    & $16\times112^2$     & R3D     & 66.5 & 29.7  \\
		IIC~\cite{tao2020selfsupervised}    & 2020   & UCF    & $16\times112^2$     & R3D     & 74.4 & 38.3  \\
		\textbf{PCL (VCOP)} & & UCF        & $16\times112^2$     & R3D     & 78.2 & 40.5  \\
		\textbf{PCL (VCP)} & & UCF         & $16\times112^2$     & R3D     & \textbf{81.1} & \textbf{45.0}  \\
	    \midrule
		VCOP~\cite{xu2019self}              & 2019   & UCF    & $16\times112^2$     & R(2+1)D & 72.4 & 30.9  \\
		VCP~\cite{luo2020video}             & 2020   & UCF    & $16\times112^2$     & R(2+1)D & 66.3 & 32.2  \\
		PRP~\cite{yao2020video}             & 2020   & UCF    & $16\times112^2$     & R(2+1)D & 72.1 & 35.0  \\
		RTT~\cite{jenni2020video}           & 2020   & UCF    & $16\times112^2$     & R(2+1)D & 81.6 & 46.4  \\
		PacePred~\cite{wang2020self}        & 2020   & UCF    & $16\times112^2$     & R(2+1)D & 75.9 & 35.9  \\
		PacePred~\cite{wang2020self}        & 2020   & K400   & $16\times112^2$     & R(2+1)D & 77.1 & 36.6  \\
		\textbf{PCL (VCOP)} & & UCF        & $16\times112^2$     & R(2+1)D     & 79.2 & 41.6  \\
		\textbf{PCL (VCP)} & & UCF         & $16\times112^2$     & R(2+1)D     & 79.9 & 45.6  \\
		\textbf{PCL (VCP)} & & K400         & $16\times112^2$     & R(2+1)D     & \textbf{85.7} & \textbf{47.4}  \\
		\midrule
		3D-RotNet~\cite{jing2018self}       & 2018   & K400   & $16\times112^2$     & R3D-18  & 62.9 & 33.7  \\
		ST-Puzzle~\cite{kim2019self}        & 2019   & K400   & $16\times112^2$     & R3D-18  & 65.8 & 33.7  \\ 
		DPC~\cite{han2019video}             & 2019   & K400   & $16\times128^2$     & R3D-18  & 68.2 & 34.5  \\
		RTT~\cite{jenni2020video}           & 2020   & UCF   & $16\times112^2$     & R3D-18  & 77.3 & 47.5  \\
		RTT~\cite{jenni2020video}           & 2020   & K400   & $16\times112^2$     & R3D-18  & 79.3 & \textbf{49.8}  \\
		\textbf{PCL (3DRotNet)} & & UCF    & $16\times112^2$     & R3D-18  & 82.8 & 47.2  \\
		\textbf{PCL (VCP)} & & UCF         & $16\times112^2$     & R3D-18  & 83.4 & 48.8  \\
		\textbf{PCL (VCP)} & & K400         & $16\times112^2$     & R3D-18  & \textbf{85.6} & 48.0  \\
		\bottomrule
		\end{tabular}
	\label{table:recog_sota}
\end{table}

\subsection{Comparison with the state-of-the-art methods}
There are too many pretext tasks in video self-supervised learning and it is impossible for us to embed our proposal to all these methods. 
The baselines we used in our study are not currently state-of-the-art methods. Some very recent works have used new pretext tasks such as pace prediction~\cite{wang2020self} or more complex temporal transformation recognition~\cite{jenni2020video} and achieved better performances. Here we compared our methods with state-of-the-art methods to demonstrate the effectiveness of our PCL. We want to clarify that there are some other works that used larger pre-trained datasets together with audio or text information of videos and achieved even higher performance~\cite{owens2018audio, korbar2018cooperative,sun2019videobert}. Here, we did not include them and only referred to these methods using similar settings for fair comparison.

The results for video retrieval in UCF101 and HMDB51 datasets are shown in Table~\ref{table:retrieval_sota} and Table~\ref{table:retrieval_hmdb}, respectively.  From Table~\ref{table:retrieval_sota}, we can see that by combining with PCL, we can easily outperform other state-of-the-art methods, no matter which backbone is used. The solution of PacePred~\cite{wang2020self} is already a combination of pretext task and contrastive learning. We can still outperform their results by a large margin based on three network backbones. The best top-1 video retrieval performance in UCF101 dataset is 55.1\%, achieved by our PCL using Resnet-18-3D network backbone and the corresponding pretext task is VCP~\cite{luo2020video}. Similar trend can be found in HMDB51 dataset in Table~\ref{table:retrieval_hmdb}. We can lift the corresponding pretext baselines by a large margin.

The results for video recognition are shown in Table~\ref{table:recog_sota}. We can observe that without our proposal, the performances of the corresponding baseline methods are lower than those of recent state-of-the-art methods. However, with the proposed PCL, which only has minor changes in the baselines, the performances can be significantly improved. In most settings, PCL performs better than the state-of-the-art methods. From this table, we can also see that the settings of existing methods vary from one to another, such as using different sizes of input data, different network architectures, and different pre-trained datasets. The total duration of Kinetics-400 dataset is around 28 days while it is about one day for UCF101 datasets. Larger datasets as well as input size will usually boost the performance. In our experiments, we only set the input size of $16\times112\times112$ while we can achieve even better performance than methods such as SpeedNet~\cite{benaim2020speednet} and MemDPC~\cite{han2020memory} when our PCL is pre-trained on UCF101 while they used larger pre-trained datasets, larger input size, and deeper networks. The best video recognition performance on UCF101 dataset is achieved when our PCL is pre-trained on Kinetics-400, reaching 85.7\%. On HMDB51, our best performance (48.8\%) is obtained by using VCP as the pretext task baseline and Resnet-18-3D network backbone, outperforming all other methods except for RTT~\cite{jenni2020video}.

It may be claimed that in some papers, their proposed pretext task or contrastive learning methods were novel and could achieve state-of-the-art performance at that time. However, based on our experiments, we find there is much room for previous methods. Exploring the limits of each method and then conducting comparison may be a fair way.

\subsection{Ablation study: effectiveness of each part}
Because we have a lot of changes on the basis of pretext tasks such as combining with contrastive learning, using residual clips, and data augmentation transformations, we want to find out how much impact each part contributes. We choose VCP as the pretext task baseline and R3D as the network backbone. Experiments are conducted on UCF101 \textit{split}~1. Results are reported in Table~\ref{tab:ablation_parts}. 
Because there are a lot of combination settings, we use the experiment ID to refer to for convenience. There are totally 16 kinds of settings for all possible situations. Here, eight out of 16 are conducted because we think it is enough to show the effectiveness of each part in our proposal.

\textbf{Residual clips.} As we can see from the comparison pair, Exp.~1 and Exp.~2, by using residual clips instead of original RGB video clips, improvements can be obtained in both video retrieval and recognition. Similar performance can be found between Exp.~5 and Exp.~7, or Exp.~6 and Exp.~8.

\textbf{Data augmentation.} We can see from Exp.~7 and Exp.~8, with strong data augmentation transformations, the top-1 performance in video retrieval can be lifted from 40.5\% to 48.1\%. And 1\% point improvement can be obtained in video recognition. From Exp.~5 and Exp.~6, we can also find that strong data augmentation is effective.

\textbf{Methods combination.} We can see a comparison set \{Exp.~1, Exp.~3, Exp.~5\}, whose experimental settings do not use our data processing strategies, a combination of VCP and contrastive learning can boost the performance of each. For comparison pair, Exp.~4 and Exp.~8, improvements can be also obtained when contrastive learning is combined with VCP in both video retrieval and recognition.  

\begin{table}[]
    \centering
    \caption{Ablation studies on different kinds of combinations. Network architecture is based on R3D. Results are reported on UCF101 \textit{split}~1. \textit{Res} means using residual clip as input and \textit{Aug} represents methods using strong data augmentations.}
    \setlength{\tabcolsep}{2mm}{
    \begin{tabular}{c c c c c c c}
    \toprule
    Exp. & Pretext & Contrastive & Res & Aug & Retrieval & Recog. \\
    \midrule
    1 & VCP      & $\times$   & $\times$   & $\times$   & 18.6 & 66.0 \\
    2 & VCP      & $\times$   & \checkmark & $\times$   & 25.6 & 77.0 \\
    3 & $\times$ & \checkmark & $\times$   & $\times$   & 34.0 & 61.2 \\
    4 & $\times$ & \checkmark & \checkmark & \checkmark & 44.7 & 79.3 \\
    5 & VCP      & \checkmark & $\times$   & $\times$   & 35.0 & 65.9 \\
    6 & VCP      & \checkmark & $\times$   & \checkmark & 40.3 & 68.9 \\
    7 & VCP      & \checkmark & \checkmark & $\times$   & 40.5 & 78.9 \\
    8 & VCP      & \checkmark & \checkmark & \checkmark & \textbf{48.1} & \textbf{79.9} \\
    \bottomrule
    \end{tabular}}
    \label{tab:ablation_parts}
\end{table}

\subsection{Ablation study: loss weight balancing}
We conducted several experiments on loss weight balancing to explore the impact of $\alpha$ in Eq.~\ref{eq:loss}. Experiments are conducted on the basis of pretext task VCP and the network backbone is R3D. Results are reported on UCF101 \textit{split}~1 in both video retrieval and recognition. 

\begin{table}[]
    \centering
    \caption{Ablation studies on the hyper-parameter $\alpha$ in Eq.~\ref{eq:loss}. Network architecture is based on R3D and the pretext task is VCP. Results are reported on UCF101 \textit{split}~1.}
    \setlength{\tabcolsep}{2mm}{
    \begin{tabular}{c c c c c c c}
    \toprule
    $\alpha$ & Top1 & Top5 & Top10 & Top20 & Top50 & Recog. \\
    \midrule
    0.1 & 43.2 & 63.1 & 72.7 & 80.9 & 89.8 & \textbf{80.1} \\
    0.5 & \textbf{48.1} & 64.7 & \textbf{73.9} & 82.0 & \textbf{90.6} & 79.9 \\
    1.0 & \textbf{48.1} & \textbf{65.8} & 73.6 & \textbf{82.1} & 90.0 & 79.3 \\
    10  & 45.9 & 65.0 & 72.7 & 81.1 & 89.6 & 73.5 \\
    \bottomrule
    \end{tabular}}
    \label{tab:ablation_alpha}
\end{table}

We can see from Table~\ref{tab:ablation_alpha}, the retrieval performances are comparable when $\alpha$ is set to 0.5 or 1.0, higher than others. However, the best recognition result is achieved when $\alpha$ is set to $0.1$. Compared with the setting $\alpha=0.1$, the top-1 retrieval accuracy is 4.9\% points higher for $\alpha=0.5$ while its corresponding recognition accuracy is 0.2\% points lower. To balance the performance in both video retrieval and recognition, we choose to set $\alpha$ to 0.5 for all of our experiments.

\section{Discussions}
In addition to the improvements on numbers, we would like to pose discussions on how a combination of pretext task and contrastive learning can yield better performance. In this section, we show some evidence and analysis towards the combination.

\subsection{General analysis}
The mechanism of pretext tasks is not well explained in theory. Researchers aim to design tasks related to their final target tasks. For example, action retrieval and action recognition require temporal information to distinguish between samples. Thus, temporal related tasks have been proposed. However, for individual pretext task, it is not clear which is the best, except based on a particular performance metrics. 

For contrastive learning, the basic idea is to distinguish one sample from another. However, determining why it functions well for motion representation extraction is difficult because spatial information may sometimes be enough. And same action clips in different instances will be treated as negatives during training.

Owing to many unclear issues, it is difficult to model the training target in a clear way. However, from the optimization target, we know that \textbf{pretext tasks focus within the sample while contrastive learning methods try to distinguish one instance from another}. By combining them together, the model can not only capture temporal information constrained by pretext tasks, but also learn discriminative features from samples constrained by contrastive learning. 

\begin{figure}[t]
    \centering
    \includegraphics[width=\columnwidth]{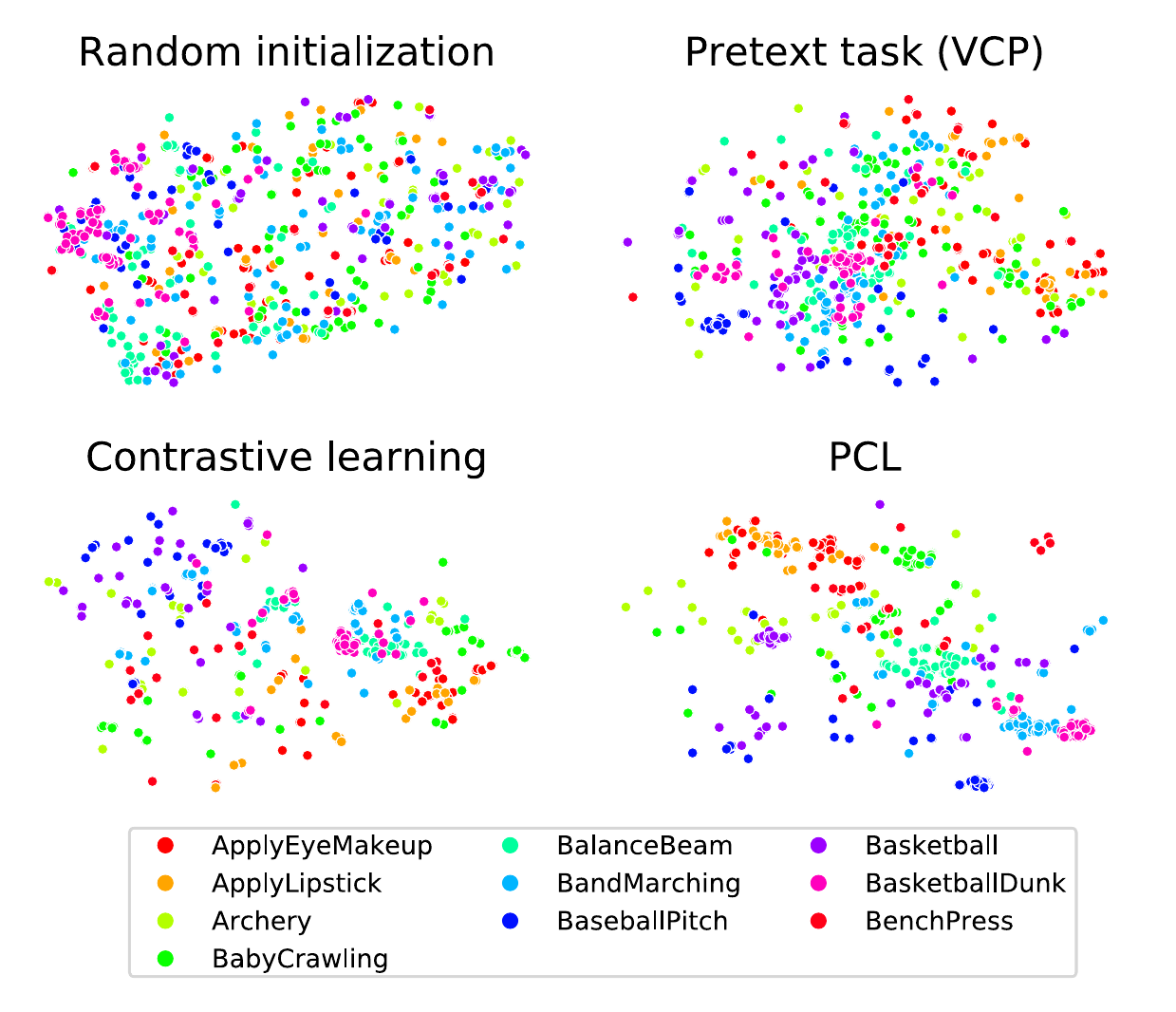} 
    \caption{Visualizations using t-SNE. The point number for PCL appears smaller because points with the same color (i.e., the same action labels) are more concentrated. The first ten categories (in alphabetical order) in UCF101 are visualized.} 
    \label{fig:tsne}
\end{figure}

\subsection{Feature visualizations}
To better understand learned features, we visualize them using t-SNE~\cite{tsne} in Fig.~\ref{fig:tsne}. Four different methods are used here: 1) random initialization, 2) one pretext task method, 3) one contrastive learning method, and 4) our proposed PCL. All models are trained in a self-supervised manner, except for the random initialization because it is initialized without training. The first ten categories in UCF101 \textit{split}~1 are visualized and each point represents one video.

As we can see from Fig.~\ref{fig:tsne}, without any training, features randomly distribute in the space. In the visualization of VCP and contrastive learning, features of the same class (in the same color) distribute more concentrated. With our PCL, it appears that the number of points is fewer because features of the same class are more close to each other and can be better clustered than the other three methods.

\subsection{Case studies}
To evaluate the advantage of pretext task, contrastive learning, and our PCL respectively, we use self-supervised trained models without changing parameters by fine-tuning. Therefore, video retrieval is used as the evaluation task. And all models are based on R3D network backbone.

\begin{figure}[t]
    \centering
    \includegraphics[width=\columnwidth]{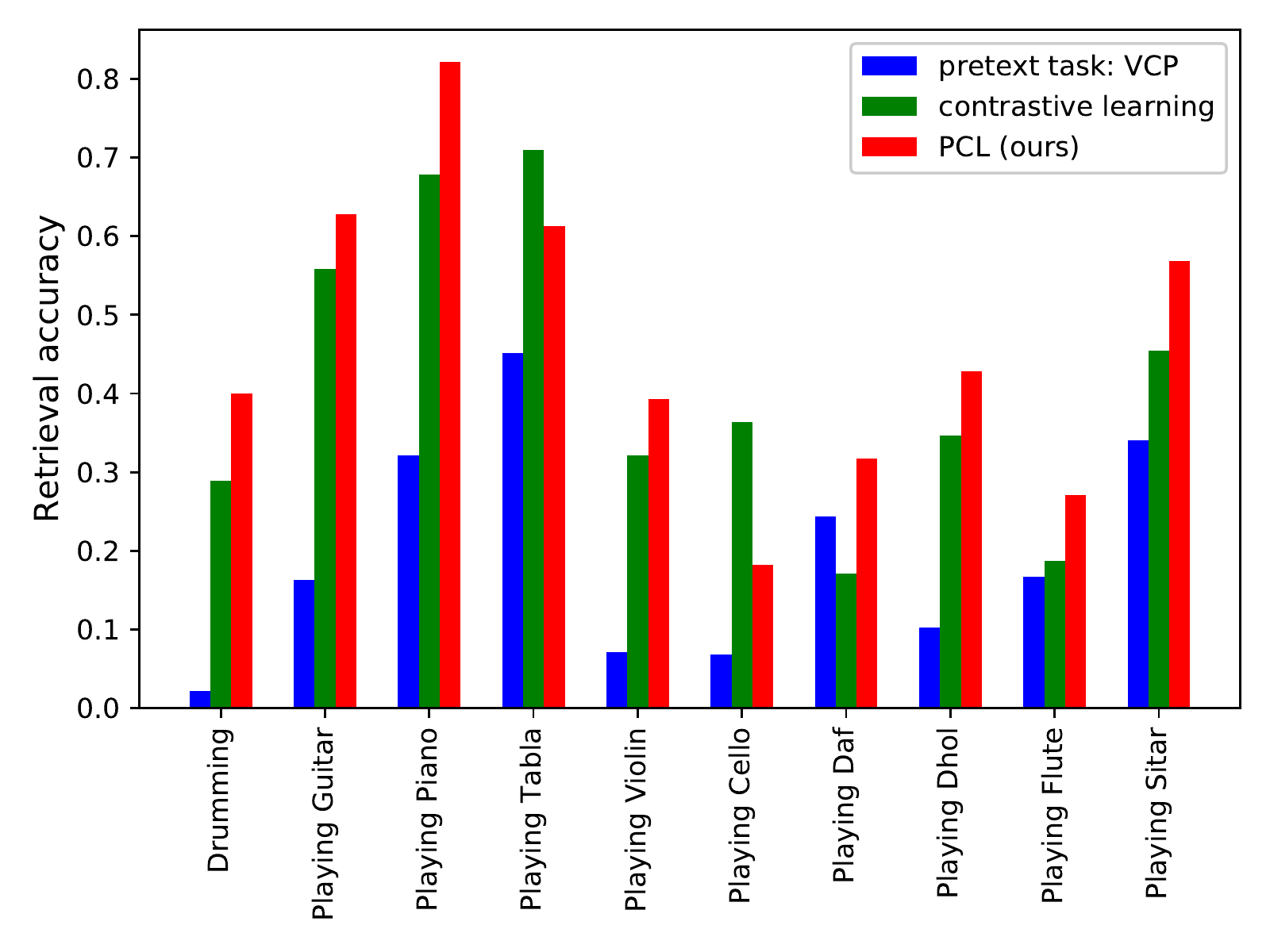} 
    \caption{Video retrieval performance on each class. All classes here belong to the \textit{Playing Musical Instruments} category. Our PCL can take the advantage of contrastive learning and compensate for pretext task baseline.} 
    \label{fig:music}
\end{figure}

One type of action categories in UCF101 dataset is \textit{Playing Musical Instruments}, where many similar actions are classified into different classes because of the different instruments. Therefore, contrastive learning should have better performance because it is constrained by distinguishing one sample from another, mainly based on spatial differences. Fig.~\ref{fig:music} illustrates this trend that contrastive learning perform better than single pretext task, VCP. Though VCP utilized temporal transformations, the movements in many cases in this category are highly similar. Because our PCL is a combination of pretext task and contrastive learning, we can see that PCL can avoid the disadvantage of pretext task and even have better performance than contrastive learning method. Note that in Fig.~\ref{fig:music}, for category \textit{Playing Tabla} and \textit{Playing Cello}, the contrastive learning method have better performance than our PCL. We find that for category \textit{Playing Tabla}, the total number of testing case is only 31, where 3 cases can cause around 10\% points decrease. For category \textit{Playing Cello}, 13.6\% testing cases for our PCL are confused with category \textit{Nunchuncks}, whose videos share similar composition with \textit{Playing Tabla}. Though it is hard to say our PCL is the best for all cases, we can still say that our combination is generally better than pretext task 
or contrastive learning methods when they stand alone.

\begin{figure}[t]
    \centering
    \includegraphics[width=\columnwidth]{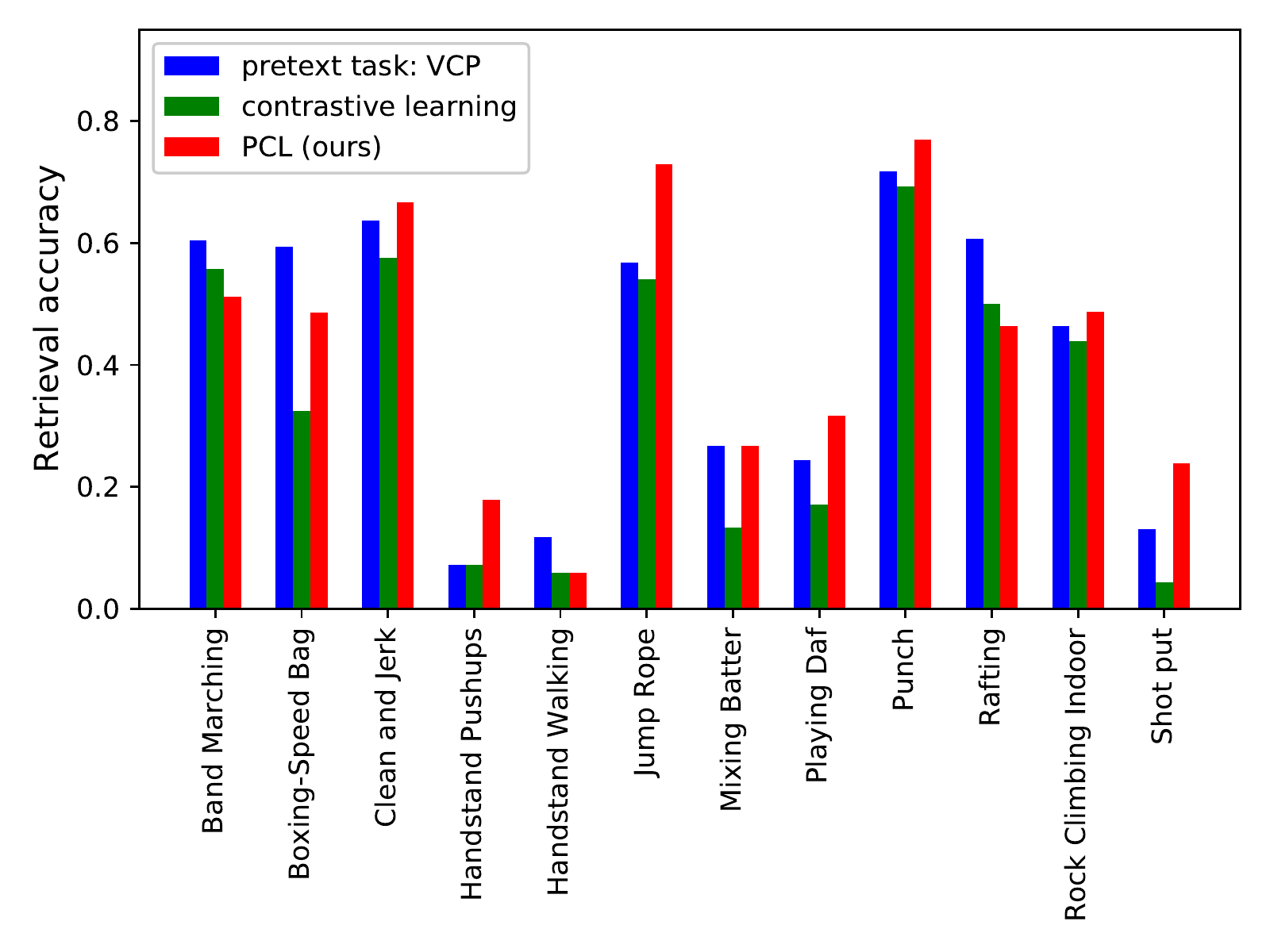} 
    \caption{Video retrieval performance on each class. The classes here are those where pretext task method perform better than contrastive learning method.Our PCL can take the advantage of pretext task baseline and compensate for contrastive learning baseline.} 
    \label{fig:pretext_better}
\end{figure}

\begin{figure}[t]
    \centering
    \includegraphics[width=\columnwidth]{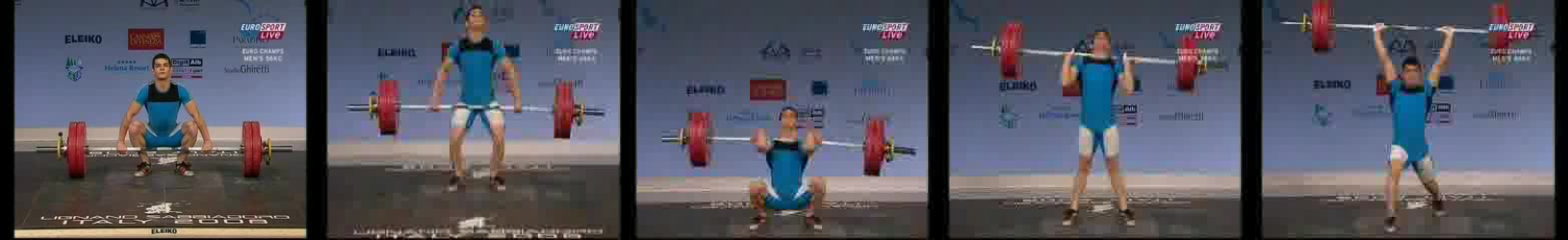} 
    \caption{Sample frames for action category \textit{Clean and Jerk}, extracted from v\_CleanAndJerk\_g11\_c03.avi in UCF101 dataset.} 
    \label{fig:cleanandjerk}
\end{figure}

Though from Table~\ref{tab:ablation_parts}, we can see that contrastive learning method can have much better performance than pretext task (44.7\% vs 25.6\%), there are still some advantage that pretext task has over contrastive learning. We visualize these classes where pretext task method perform better than contrastive learning. As we can see from Fig.~\ref{fig:pretext_better}, some of these categories are much more temporal related. For example, the action \textit{Clean and Jerk} represents a series of movements (Fig.~\ref{fig:cleanandjerk}). Because the pretext task VCP contain temporal transformations, it enable the model to capture more temporal related features. And we can also see from the histogram that our PCL can achieve the better or comparable performance.

In the tiny experiments, our PCL can achieve the best performance in 72 out of 101 classes, and the best averaging results, revealing that a combination of pretext task and contrastive learning can take the advantage of both.

\section{Conclusions}
In this paper, we proposed Pretext-Contrastive Learning (PCL), a joint optimization framework facilitating both pretext tasks and contrastive learning, which is beyond a simple combination. Data processing strategies such as residual clips and strong data augmentations are used in our framework. Extensive ablation studies showed the effectiveness of each component in our proposal. Experiments using different pretext task baselines with different network backbones in different evaluation tasks on two benchmark datasets revealed the effectiveness and the generality of our proposal. With our PCL framework and the empirical settings, pretext tasks and contrastive learning can boost each other, and old benchmarking baselines can be lifted to a new level, which could provide a guideline for the self-supervised video representation community. Our proposed PCL is sufficiently flexible enough and can be easily applied to almost any existing pretext task or contrastive method.

\section*{Acknowledgments}
This work was partially financially supported by the Grants-in-Aid for Scientific Research Numbers JP19K20289 and JP18H03339 from JSPS.


\ifCLASSOPTIONcaptionsoff
  \newpage
\fi



%
\bibliographystyle{IEEEtran}
\bibliography{egbib}

\begin{thebibliography}{10}
\providecommand{\url}[1]{#1}
\csname url@samestyle\endcsname
\providecommand{\newblock}{\relax}
\providecommand{\bibinfo}[2]{#2}
\providecommand{\BIBentrySTDinterwordspacing}{\spaceskip=0pt\relax}
\providecommand{\BIBentryALTinterwordstretchfactor}{4}
\providecommand{\BIBentryALTinterwordspacing}{\spaceskip=\fontdimen2\font plus
\BIBentryALTinterwordstretchfactor\fontdimen3\font minus
  \fontdimen4\font\relax}
\providecommand{\BIBforeignlanguage}[2]{{%
\expandafter\ifx\csname l@#1\endcsname\relax
\typeout{** WARNING: IEEEtran.bst: No hyphenation pattern has been}%
\typeout{** loaded for the language `#1'. Using the pattern for}%
\typeout{** the default language instead.}%
\else
\language=\csname l@#1\endcsname
\fi
#2}}
\providecommand{\BIBdecl}{\relax}
\BIBdecl

\bibitem{krizhevsky2012imagenet}
A.~Krizhevsky, I.~Sutskever, and G.~E. Hinton, ``Imagenet classification with
  deep convolutional neural networks,'' in \emph{Advances in neural information
  processing systems}, 2012, pp. 1097--1105.

\bibitem{kinetics}
J.~{Carreira} and A.~{Zisserman}, ``Quo vadis, action recognition? a new model
  and the kinetics dataset,'' pp. 4724--4733, 2017.

\bibitem{luo2020video}
D.~Luo, C.~Liu, Y.~Zhou, D.~Yang, C.~Ma, Q.~Ye, and W.~Wang, ``Video cloze
  procedure for self-supervised spatio-temporal learning,'' in
  \emph{Proceedings of the AAAI Conference on Artificial Intelligence},
  vol.~34, no.~07, 2020, pp. 11\,701--11\,708.

\bibitem{noroozi2016unsupervised}
M.~Noroozi and P.~Favaro, ``Unsupervised learning of visual representations by
  solving jigsaw puzzles,'' in \emph{European Conference on Computer Vision},
  2016, pp. 69--84.

\bibitem{pathak2016context}
D.~Pathak, P.~Krahenbuhl, J.~Donahue, T.~Darrell, and A.~A. Efros, ``Context
  encoders: Feature learning by inpainting,'' in \emph{Proceedings of the
  IEEE/CVF Conference on Computer Vision and Pattern Recognition}, 2016, pp.
  2536--2544.

\bibitem{gidaris2018unsupervised}
S.~Gidaris, P.~Singh, and N.~Komodakis, ``Unsupervised representation learning
  by predicting image rotations,'' \emph{International conference on learning
  representations}, 2018.

\bibitem{misra2016shuffle}
I.~Misra, C.~L. Zitnick, and M.~Hebert, ``Shuffle and learn: unsupervised
  learning using temporal order verification,'' in \emph{European Conference on
  Computer Vision}, 2016, pp. 527--544.

\bibitem{fernando2017self}
B.~Fernando, H.~Bilen, E.~Gavves, and S.~Gould, ``Self-supervised video
  representation learning with odd-one-out networks,'' in \emph{Proceedings of
  the IEEE/CVF Conference on Computer Vision and Pattern Recognition}, 2017,
  pp. 3636--3645.

\bibitem{lee2017unsupervised}
H.-Y. Lee, J.-B. Huang, M.~Singh, and M.-H. Yang, ``Unsupervised representation
  learning by sorting sequences,'' in \emph{IEEE International Conference on
  Computer Vision}, 2017, pp. 667--676.

\bibitem{xu2019self}
D.~Xu, J.~Xiao, Z.~Zhao, J.~Shao, D.~Xie, and Y.~Zhuang, ``Self-supervised
  spatiotemporal learning via video clip order prediction,'' in
  \emph{Proceedings of the IEEE/CVF Conference on Computer Vision and Pattern
  Recognition}, 2019, pp. 10\,334--10\,343.

\bibitem{benaim2020speednet}
S.~Benaim, A.~Ephrat, O.~Lang, I.~Mosseri, W.~T. Freeman, M.~Rubinstein,
  M.~Irani, and T.~Dekel, ``Speednet: Learning the speediness in videos,'' in
  \emph{Proceedings of the IEEE/CVF Conference on Computer Vision and Pattern
  Recognition}, 2020, pp. 9922--9931.

\bibitem{cho2020self}
H.~Cho, T.~Kim, H.~J. Chang, and W.~Hwang, ``Self-supervised spatio-temporal
  representation learning using variable playback speed prediction,''
  \emph{arXiv preprint arXiv:2003.02692}, 2020.

\bibitem{yao2020video}
Y.~Yao, C.~Liu, D.~Luo, Y.~Zhou, and Q.~Ye, ``Video playback rate perception
  for self-supervised spatio-temporal representation learning,'' in
  \emph{Proceedings of the IEEE/CVF Conference on Computer Vision and Pattern
  Recognition}, 2020, pp. 6548--6557.

\bibitem{jenni2020video}
S.~Jenni, G.~Meishvili, and P.~Favaro, ``Video representation learning by
  recognizing temporal transformations,'' \emph{European Conference on Computer
  Vision}, 2020.

\bibitem{wang2020self}
J.~Wang, J.~Jiao, and Y.-H. Liu, ``Self-supervised video representation
  learning by pace prediction,'' \emph{European Conference on Computer Vision},
  2020.

\bibitem{piergiovanni2020evolving}
A.~Piergiovanni, A.~Angelova, and M.~S. Ryoo, ``Evolving losses for
  unsupervised video representation learning,'' in \emph{Proceedings of the
  IEEE/CVF Conference on Computer Vision and Pattern Recognition}, 2020, pp.
  133--142.

\bibitem{chen2020simple}
T.~Chen, S.~Kornblith, M.~Norouzi, and G.~Hinton, ``A simple framework for
  contrastive learning of visual representations,'' \emph{ICML}, 2020.

\bibitem{he2020momentum}
K.~He, H.~Fan, Y.~Wu, S.~Xie, and R.~Girshick, ``Momentum contrast for
  unsupervised visual representation learning,'' in \emph{Proceedings of the
  IEEE/CVF Conference on Computer Vision and Pattern Recognition}, 2020, pp.
  9729--9738.

\bibitem{wu2018unsupervised}
Z.~Wu, Y.~Xiong, S.~X. Yu, and D.~Lin, ``Unsupervised feature learning via
  non-parametric instance discrimination,'' in \emph{Proceedings of the
  IEEE/CVF Conference on Computer Vision and Pattern Recognition}, 2018, pp.
  3733--3742.

\bibitem{oord2018representation}
A.~v.~d. Oord, Y.~Li, and O.~Vinyals, ``Representation learning with
  contrastive predictive coding,'' \emph{arXiv preprint arXiv:1807.03748},
  2018.

\bibitem{tian2019contrastive}
Y.~Tian, D.~Krishnan, and P.~Isola, ``Contrastive multiview coding,''
  \emph{European Conference on Computer Vision}, 2020.

\bibitem{tao2020rethinking}
L.~Tao, X.~Wang, and T.~Yamasaki, ``Rethinking motion representation: Residual
  frames with 3d convnets for better action recognition,'' \emph{arXiv preprint
  arXiv:2001.05661}, 2020.

\bibitem{tao2020motion}
L.~{Tao}, X.~Wang, and T.~Yamasaki, ``Motion representation using residual
  frames with 3d cnn,'' \emph{IEEE International Conference on Image
  Processing}, pp. 1786--1790, 2020.

\bibitem{zhang2016colorful}
R.~Zhang, P.~Isola, and A.~A. Efros, ``Colorful image colorization,'' in
  \emph{European Conference on Computer Vision}, 2016, pp. 649--666.

\bibitem{hinton2006reducing}
G.~E. Hinton and R.~R. Salakhutdinov, ``Reducing the dimensionality of data
  with neural networks,'' \emph{Science}, vol. 313, no. 5786, pp. 504--507,
  2006.

\bibitem{kingma2013auto}
D.~P. Kingma and M.~Welling, ``Auto-encoding variational bayes,''
  \emph{International conference on learning representations}, 2014.

\bibitem{jing2018self}
L.~Jing, X.~Yang, J.~Liu, and Y.~Tian, ``Self-supervised spatiotemporal feature
  learning via video rotation prediction,'' \emph{arXiv preprint
  arXiv:1811.11387}, 2018.

\bibitem{kim2019self}
D.~Kim, D.~Cho, and I.~S. Kweon, ``Self-supervised video representation
  learning with space-time cubic puzzles,'' in \emph{Proceedings of the AAAI
  Conference on Artificial Intelligence}, vol.~33, 2019, pp. 8545--8552.

\bibitem{Jing2020pami}
L.~{Jing} and Y.~{Tian}, ``Self-supervised visual feature learning with deep
  neural networks: A survey,'' \emph{IEEE Transactions on Pattern Analysis and
  Machine Intelligence}, pp. 1--1, 2020.

\bibitem{hadsell2006dimensionality}
R.~Hadsell, S.~Chopra, and Y.~LeCun, ``Dimensionality reduction by learning an
  invariant mapping,'' in \emph{Proceedings of the IEEE/CVF Conference on
  Computer Vision and Pattern Recognition}, vol.~2, 2006, pp. 1735--1742.

\bibitem{hjelm2018learning}
R.~D. Hjelm, A.~Fedorov, S.~Lavoie-Marchildon, K.~Grewal, P.~Bachman,
  A.~Trischler, and Y.~Bengio, ``Learning deep representations by mutual
  information estimation and maximization,'' \emph{International conference on
  learning representations}, 2019.

\bibitem{chen2020improved}
X.~Chen, H.~Fan, R.~Girshick, and K.~He, ``Improved baselines with momentum
  contrastive learning,'' \emph{arXiv preprint arXiv:2003.04297}, 2020.

\bibitem{grill2020bootstrap}
J.-B. Grill, F.~Strub, F.~Altch{\'e}, C.~Tallec, P.~H. Richemond,
  E.~Buchatskaya, C.~Doersch, B.~A. Pires, Z.~D. Guo, M.~G. Azar \emph{et~al.},
  ``Bootstrap your own latent: A new approach to self-supervised learning,''
  \emph{Advances in Neural Information Processing Systems}, 2020.

\bibitem{han2019video}
T.~{Han}, W.~{Xie}, and A.~{Zisserman}, ``Video representation learning by
  dense predictive coding,'' in \emph{IEEE/CVF International Conference on
  Computer Vision Workshop}, 2019, pp. 1483--1492.

\bibitem{han2020memory}
T.~Han, W.~Xie, and A.~Zisserman, ``Memory-augmented dense predictive coding
  for video representation learning,'' \emph{European Conference on Computer
  Vision}, 2020.

\bibitem{tao2020selfsupervised}
L.~Tao, X.~Wang, and T.~Yamasaki, ``Self-supervised video representation
  learning using inter-intra contrastive framework,'' in \emph{Proceedings of
  the 28th ACM International Conference on Multimedia}, 2020, pp. 2193--2201.

\bibitem{owens2018audio}
A.~Owens and A.~A. Efros, ``Audio-visual scene analysis with self-supervised
  multisensory features,'' in \emph{European Conference on Computer Vision},
  2018, pp. 631--648.

\bibitem{korbar2018cooperative}
B.~Korbar, D.~Tran, and L.~Torresani, ``Cooperative learning of audio and video
  models from self-supervised synchronization,'' in \emph{Advances in Neural
  Information Processing Systems}, 2018, pp. 7763--7774.

\bibitem{sun2019videobert}
C.~Sun, A.~Myers, C.~Vondrick, K.~Murphy, and C.~Schmid, ``Videobert: A joint
  model for video and language representation learning,'' in \emph{IEEE
  International Conference on Computer Vision}, 2019, pp. 7464--7473.

\bibitem{gutmann2010noise}
M.~Gutmann and A.~Hyv{\"a}rinen, ``Noise-contrastive estimation: A new
  estimation principle for unnormalized statistical models,'' in
  \emph{Proceedings of the Thirteenth International Conference on Artificial
  Intelligence and Statistics}, 2010, pp. 297--304.

\bibitem{wang2020hypersphere}
T.~Wang and P.~Isola, ``Understanding contrastive representation learning
  through alignment and uniformity on the hypersphere,'' in \emph{International
  Conference on Machine Learning}, 2020, pp. 9929--9939.

\bibitem{ucf101}
K.~Soomro, A.~R. Zamir, and M.~Shah, ``Ucf101: A dataset of 101 human actions
  classes from videos in the wild,'' \emph{arXiv preprint arXiv:1212.0402},
  2012.

\bibitem{hmdb}
H.~{Kuehne}, H.~{Jhuang}, E.~{Garrote}, T.~{Poggio}, and T.~{Serre}, ``Hmdb: A
  large video database for human motion recognition,'' in \emph{IEEE
  International Conference on Computer Vision}, 2011, pp. 2556--2563.

\bibitem{i3d}
J.~Carreira and A.~Zisserman, ``Quo vadis, action recognition? a new model and
  the kinetics dataset,'' in \emph{Proceedings of the IEEE/CVF Conference on
  Computer Vision and Pattern Recognition}, 2017, pp. 6299--6308.

\bibitem{c3d}
D.~{Tran}, L.~{Bourdev}, R.~{Fergus}, L.~{Torresani}, and M.~{Paluri},
  ``Learning spatiotemporal features with 3d convolutional networks,'' in
  \emph{IEEE International Conference on Computer Vision}, 2015, pp.
  4489--4497.

\bibitem{res3d}
K.~{Hara}, H.~{Kataoka}, and Y.~{Satoh}, ``Can spatiotemporal 3d cnns retrace
  the history of 2d cnns and imagenet?'' in \emph{Proceedings of the IEEE/CVF
  Conference on Computer Vision and Pattern Recognition}, 2018, pp. 6546--6555.

\bibitem{r3d}
D.~{Tran}, H.~{Wang}, L.~{Torresani}, J.~{Ray}, Y.~{LeCun}, and M.~{Paluri},
  ``A closer look at spatiotemporal convolutions for action recognition,'' in
  \emph{Proceedings of the IEEE/CVF Conference on Computer Vision and Pattern
  Recognition}, 2018, pp. 6450--6459.

\bibitem{resnet}
K.~{He}, X.~{Zhang}, S.~{Ren}, and J.~{Sun}, ``Deep residual learning for image
  recognition,'' in \emph{Proceedings of the IEEE/CVF Conference on Computer
  Vision and Pattern Recognition}, 2016, pp. 770--778.

\bibitem{s3d}
S.~Xie, C.~Sun, J.~Huang, Z.~Tu, and K.~Murphy, ``Rethinking spatiotemporal
  feature learning: Speed-accuracy trade-offs in video classification,'' in
  \emph{European Conference on Computer Vision}, 2018, pp. 305--321.

\bibitem{tsne}
L.~v.~d. Maaten and G.~Hinton, ``Visualizing data using t-sne,'' \emph{Journal
  of machine learning research}, vol.~9, no. Nov, pp. 2579--2605, 2008.

\end{thebibliography}
%

\end{document}